\documentclass{IET-Conf-Paper}
\usepackage{booktabs}
\usepackage{graphicx}
\usepackage{subcaption}
\usepackage{acronym}
\usepackage{hyperref}
\usepackage{lmodern}

\begin{document}
% ------------------------------------------------------------
% Acronyms for DPSP 2026 paper
% ------------------------------------------------------------

% --- General / Power Systems ---
\newacro{ac}[AC]{alternating current}
\newacro{dc}[DC]{direct current}
\newacro{hv}[HV]{high voltage}
\newacro{mv}[MV]{medium voltage}
\newacro{lv}[LV]{low voltage}
\newacro{hvdc}[HVDC]{high-voltage direct current}
\newacro{der}[DER]{distributed energy resource}
\newacro{scada}[SCADA]{supervisory control and data acquisition}
\newacro{ied}[IED]{intelligent electronic device}
\newacro{pmu}[PMU]{phasor measurement unit}
\newacro{pr}[PR]{protection relay}
\newacro{rtds}[RTDS]{real-time digital simulator}
\newacro{emt}[EMT]{electromagnetic transients}
\newacro{sip}[SIP]{system integrity protection}
\newacro{iec}[IEC]{International Electrotechnical Commission}

% --- Fault & Protection Tasks ---
\newacro{fd}[FD]{fault detection}
\newacro{fc}[FC]{fault classification}
\newacro{fli}[FLI]{fault line identification}
\newacro{fl}[FL]{fault localization}
\newacro{mae}[MAE]{mean absolute error}
\newacro{mape}[MAPE]{mean absolute percentage error}
\newacro{rmse}[RMSE]{root mean square error}

% --- Machine Learning / Deep Learning ---
\newacro{ai}[AI]{artificial intelligence}
\newacro{ml}[ML]{machine learning}
\newacro{dl}[DL]{deep learning}
\newacro{cnn}[CNN]{convolutional neural network}
\newacro{rnn}[RNN]{recurrent neural network}
\newacro{lstm}[LSTM]{long short-term memory}
\newacro{gru}[GRU]{gated recurrent unit}
\newacro{tcn}[TCN]{temporal convolutional network}
\newacro{tft}[TFT]{temporal fusion transformer}
\newacro{mlp}[MLP]{multilayer perceptron}
\newacro{drl}[DRL]{deep reinforcement learning}
\newacro{kd}[KD]{knowledge distillation}

% --- Datasets & Metrics ---
\newacro{f1}[F1]{F1-score}
\newacro{r2}[$R^2$]{coefficient of determination}
\newacro{roc}[ROC]{receiver operating characteristic}
\newacro{auc}[AUC]{area under the curve}
\newacro{cv}[CV]{cross-validation}

% --- Tools / Frameworks ---
\newacro{hpc}[HPC]{high-performance computing}
\newacro{gpu}[GPU]{graphics processing unit}
\newacro{mlflow}[MLflow]{machine learning flow}
\newacro{yaml}[YAML]{YAML ain’t markup language}
\newacro{api}[API]{application programming interface}
\newacro{csv}[CSV]{comma-separated values}

% --- Signal Processing & Representations ---
\newacro{fft}[FFT]{fast Fourier transform}
\newacro{stft}[STFT]{short-time Fourier transform}
\newacro{cwt}[CWT]{continuous wavelet transform}
\newacro{tf}[TF]{time–frequency}
\newacro{gaf}[GAF]{Gramian angular field}

% --- Institutions / Venues ---
\newacro{pscc}[PSCC]{Power Systems Computation Conference}
\newacro{dpsp}[DPSP]{Developments in Power System Protection}
\newacro{fau}[FAU]{Friedrich-Alexander-Universität Erlangen–Nürnberg}
\newacro{prl}[PRL]{Pattern Recognition Lab}
\newacro{iees}[IEES]{Institute of Electrical Energy Systems}

\title{ROBUSTNESS EVALUATION OF MACHINE LEARNING MODELS FOR FAULT CLASSIFICATION AND LOCALIZATION IN POWER SYSTEM PROTECTION}

\author{Julian Oelhaf\ad{1}\textsuperscript{*}, Mehran Pashaei\ad{1}, Georg Kordowich\ad{2}, Christian Bergler\ad{3}, Andreas Maier\ad{1}, Johann J{\"a}ger\ad{2}, Siming Bayer\ad{1}}

\address{\add{1}{Pattern Recognition Lab, Friedrich-Alexander-Universität Erlangen-Nürnberg, Germany} \add{2}{Institute of Electrical Energy Systems, Friedrich-Alexander-Universität Erlangen-Nürnberg, Germany}
\add{3}{Department of Electrical Engineering, Media and Computer Science, Ostbayerische Technische Hochschule Amberg-Weiden, Germany}
\email{julian.oelhaf@fau.de}
}


\begin{abstract} % 200 words
The growing penetration of renewable and distributed generation is transforming power systems and challenging conventional protection schemes that rely on fixed settings and local measurements.  
Machine learning (ML) offers a data-driven alternative for centralized fault classification (FC) and fault localization (FL), enabling faster and more adaptive decision-making.  
However, practical deployment critically depends on robustness. Protection algorithms must remain reliable even when confronted with missing, noisy, or degraded sensor data.  
This work introduces a unified framework for systematically evaluating the robustness of ML models in power system protection.
High-fidelity EMT simulations are used to model realistic degradation scenarios, including sensor outages, reduced sampling rates, and transient communication losses.  
The framework provides a consistent methodology for benchmarking models, quantifying the impact of limited observability, and identifying critical measurement channels required for resilient operation.  
Results show that FC remains highly stable under most degradation types but drops by about 13\,\% under single-phase loss, while FL is more sensitive overall, with current loss increasing localization error by over 150\,\%.
These findings offer actionable guidance for robustness-aware design of future ML-assisted protection systems.
\end{abstract}

\maketitle
\begingroup
\renewcommand\thefootnote{}\footnotetext{
\scriptsize
This paper is a postprint of a paper submitted to and accepted for publication in the
20th IET International Conference on Developments in Power System Protection (DPSP Global 2026)
and is subject to Institution of Engineering and Technology Copyright.
The copy of record is available at the IET Digital Library.
}
\endgroup

\section{Introduction}
Modern power grids are undergoing a fundamental transformation driven by the large-scale integration of renewable and decentralized energy sources. This transition increases system complexity, leading to dynamic power flows and diverse fault characteristics that challenge conventional protection schemes~\cite{telukunta_protection_2017}.
While traditional methods remain effective for rapid fault isolation, they lack the analytical depth required for advanced post-fault tasks such as \ac{fc} and \ac{fl}, which are essential for efficient grid restoration. 

\Ac{ml} offers a promising alternative for developing centralized, data-driven protection systems capable of deeper and more adaptive fault analysis beyond simple detection.  
However, their successful and reliable deployment largely depends on data quality, as real-world systems usually operate under imperfect conditions caused by sensor failures, communication issues, or hardware limitations.  
Reliable and robust operation under partial observability is therefore a key requirement for practical \ac{ml}-based protection.

Our recent scoping review~\cite{oelhaf_scoping_2025} provides a comprehensive overview of \ac{ml} applications in power system protection and highlights major inconsistencies in datasets, preprocessing strategies, and evaluation metrics that hinder comparability.  
Most existing studies investigate \ac{fc} or \ac{fl} separately and under ideal measurement conditions, with only a few considering the impact of measurement noise or missing data.  
Although recent works across HVDC systems, wind farms, and hybrid grids~\cite{da_silva_intelligent_2025,kandil_enhancing_2025,mishra_deep_2025,vaidya_fault_2025,yadav_integrating_2025} demonstrate methodological advances, they remain fragmented and rarely evaluate robustness.  
To the best of our knowledge, no prior study has systematically examined how \ac{ml}-based \ac{fc} and \ac{fl} perform under degraded or incomplete data.

Building on our previous framework for fault detection and line identification~\cite{oelhaf_impact_2025}, this study extends the analysis to a unified robustness evaluation across both classification and localization tasks.  
The proposed framework systematically models realistic degradation scenarios, including sensor outages, reduced sampling rates, and communication losses, to quantify their impact on model performance.  
It establishes a consistent benchmark and identifies the critical measurement channels and operating conditions required for reliable and resilient \ac{ml}-assisted power system protection.

\section{Methodology}\label{sec:methodology}

In order to evaluate the robustness of \ac{ml} models for \ac{fc} and \ac{fl}, a structured framework has been designed, comprising four stages: data generation, preprocessing, model training, and controlled introduction of measurement limitations to emulate real-world conditions. Model performance is quantified using well-defined evaluation metrics.

\subsection{Data Material}\label{sec:dataset}

\begin{figure*}[!htbp]
	\centering
	\includegraphics[trim={0 0 0 0},clip,width=0.9\linewidth]{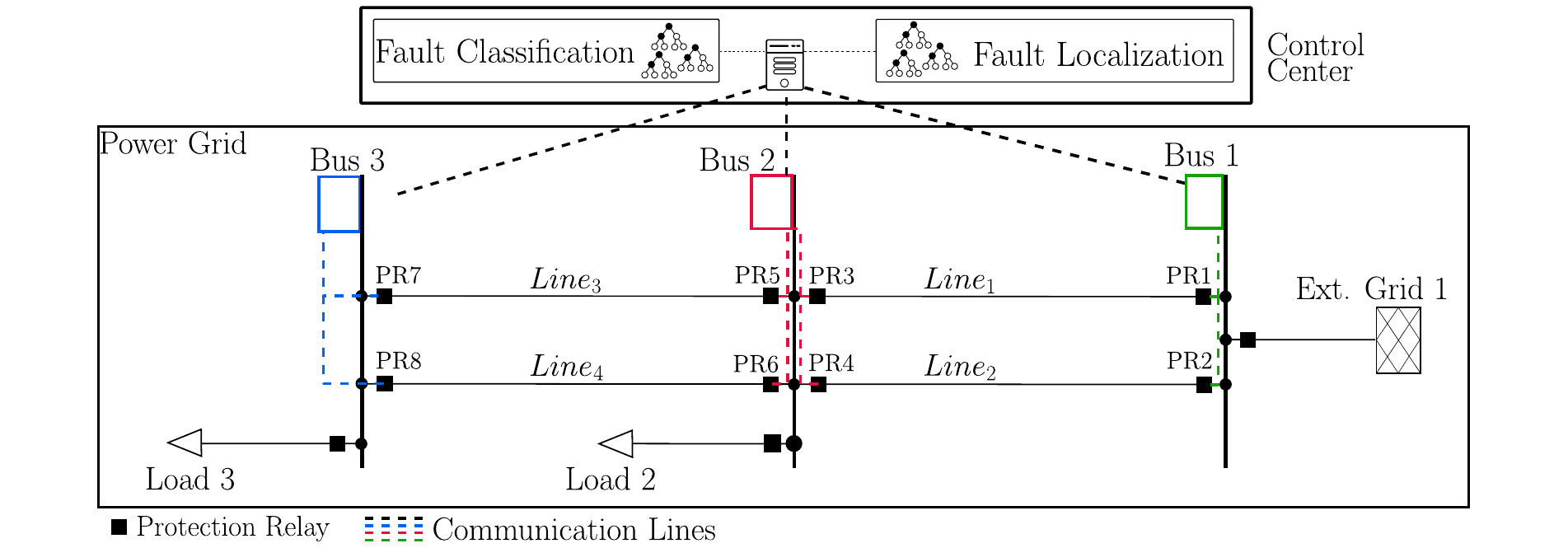}
    \caption{Schematic of the ``Double Line'' grid topology showing transmission lines, \acp{pr}, and communication links. Measurements from each \ac{pr} are transmitted to the corresponding substation and then forwarded to the control centre for \ac{fc} and \ac{fl}.}
    \label{fig:double_line_grid_topology}
    % \vspace{-2mm} % Adjust vertical spacing if necessary
\end{figure*}

The dataset is generated following the procedure of Wang et~al.~\cite{wang_generic_2022} using DIgSILENT PowerFactory\footnote{\href{https://www.digsilent.de/en/}{https://www.digsilent.de/en/}} for \ac{emt} simulations of the benchmark double line grid~\cite{meyer_hybrid_2020} shown in Figure~\ref{fig:double_line_grid_topology}. \ac{emt} modelling provides high-resolution voltage and current waveforms that accurately capture transient fault behaviour~\cite{mahr_distanzschutzalgorithmen_2021}.  
To enhance data diversity and model generalization, domain randomization is applied by varying line lengths, load conditions, fault locations, and external grid parameters within realistic operating ranges~\cite{roeper_kurzschlusstrome_1984,oeding_elektrische_2016}.

The dataset consists of 9{,}022 simulated fault cases, each lasting 1\,s at a nominal voltage of 90\,kV and sampled at 6{,}400\,Hz.  
It includes single-, double-, double-to-ground, and three-phase short circuits (denoted by the faulted phases A, B, and C). 
Each transmission line is monitored by one \ac{pr} per terminal, resulting in eight \acp{pr} and 48 total channels recording three-phase currents and voltages:
\[
\begin{aligned}
    I_{PR}(t) &= (I_A(t), I_B(t), I_C(t)), \\
    V_{PR}(t) &= (V_A(t), V_B(t), V_C(t)),
    \quad t \in [0,1]\,\text{s}.
\end{aligned}
\]

Combining all sensors yields a multivariate time series for each simulation case~$n$: 
\[
X^{(n)} = [x^{(n)}_1, x^{(n)}_2, \ldots, x^{(n)}_T] \in \mathbb{R}^{48 \times T},
\quad x^{(n)}_t \in \mathbb{R}^{48},
\]
where $T$ (=\,6{,}400) denotes the total number of time steps per simulation, and $n$ indexes the individual fault cases.

\subsection{Data Preprocessing}
Each episode is cropped to $\pm80$\,ms around the fault inception and segmented into overlapping windows of duration $w = 50$\,ms with a stride of 5\,ms following~\cite{oelhaf_systematic_2025}.  
At a sampling rate of 6.4\,kHz, each window has a length of $L_w = w \times f_s = 320$ samples across 48 channels, forming an input tensor $X^{(n)}_i \in \mathbb{R}^{L_w \times 48}$ (15{,}360 data points).  
This segmentation yields 207{,}506 windows in total, of which 81{,}119 contain fault events (at least partially).  
The high simulation frequency ensures realistic temporal resolution, matching real-time protection data and enabling controlled downsampling experiments.

\subsection{Machine Learning for Fault Classification and Localization}

The preprocessed input tensors $X^{(n)}_i \in \mathbb{R}^{L_w \times 48}$ serve as inputs for two protection tasks: \ac{fc} and \ac{fl}, following the methodology proposed in~\cite{oelhaf_benchmarking_2025}.  
For \ac{fc}, each window $X^{(n)}_i$ is assigned one of eleven labels, ten short-circuit types plus a no-fault class, and evaluated using the F1-score.  
For \ac{fl}, the same input is used to predict a continuous target $y_{\text{FL}} \in [0.01,\,0.99]$, representing the fault location as a percentage of the line length.  
This regression task mirrors distance protection principles~\cite{horowitz_power_2023} and is evaluated using the \ac{mae} (in~\% of line length) as metric.

Both tasks use as a model architecture a compact \ac{mlp} with two hidden layers and ReLU activations, identified in our benchmark study~\cite{oelhaf_benchmarking_2025} as the best-performing and efficient \ac{ml} model across tasks.  
Each model maps the flattened input window $X^{(n)}_i$ through fully connected layers to a latent representation.  
The output layer contains 11 softmax-activated neurons for \ac{fc} and a single linear neuron for \ac{fl} to estimate the fault location.

All models were trained using the the Adam optimizer together with an initial learning rate of $1.38\times10^{-4}$ and an early stopping criterion of 20 epochs without any improvement on the validation set comprising 10\% of the data.  
To prevent temporal leakage, windows are grouped and split at the episode level as part of a five-fold cross-validation (CV).  
A baseline model is trained on the complete dataset, and robustness is assessed by comparing relative performance changes under different data degradation scenarios.

A key part of this study is assessing model robustness under data sparsity, a common challenge in operational power systems.  
Following the procedure established in~\cite{oelhaf_impact_2025}, sparse measurement conditions were simulated by systematically degrading the original simulation data according to three realistic scenarios.

%\textbf{Sensor and Component Failure.}  
The first scenario describes modelling permanent data loss within the secondary systems, caused by hardware or communication malfunctions in the protection, control, or measurement infrastructure.
Failures at different scales were simulated, including the loss of all data from a bus (substation), individual \acp{pr}, or specific sensor types, such as all voltage,  current, or single-phase measurements (e.g., Phase~A).
%\textbf{Reduced Sampling Rate:}  
The second scenario illustrates reduced sampling rates, in order to represent constraints imposed by legacy hardware or limited bandwidth. The original 6.4\,kHz data has been downsampled to frequencies as low as 100\,Hz, corresponding to maximal reduction factor of $\times$64.
The third and last scenario addresses temporal communication loss. Short-term dropouts were simulated by zeroing contiguous 5-40\,ms segments across all input features, emulating transient data losses in synchronized measurements.

\begin{table}[tbp]
\centering
\caption{Overview of data sparsity scenarios.}
\renewcommand{\arraystretch}{1.01}
\resizebox{0.98\linewidth}{!}{
\begin{tabular}{@{}ll@{}}
\toprule
\textbf{Sparsity Scenario} & \textbf{Parameter Values} \\
\midrule
Missing Voltage & True, False \\
Missing Current & True, False \\
Reduced Sampling Rate & 6.4\,kHz, 3.2\,kHz, 1.6\,kHz  \\
                    & 800\,Hz, 400\,Hz, 200\,Hz, 100\,Hz \\
Relay Comm Failure & 1 - 8 \\
Substation Comm. Failure & 1, 2, 3 \\
Phase Measurement Failure & A, B, C \\
Temporal Comm Loss & 5 - 40\,ms \\
\bottomrule
\end{tabular}}
\label{tab:scenarios}
\end{table}

Table~\ref{tab:scenarios} summarizes the different data degradation scenarios and their corresponding parameter variations. Each scenario introduces a unique challenge for the \ac{fc}  and \ac{fl} models, helping us to evaluate their robustness under realistic conditions of limited observability. This systematic assessment is crucial for understanding the resilience of ML-based fault analysis in real-world applications, where imperfect data is a common operational reality.

\section{Experiments and Results}  
Model robustness was evaluated under the data degradation scenarios described in Section~\ref{sec:methodology}. Thus, performance was measured using the F1-score for \ac{fc} and the \ac{mae}  for \ac{fl}, relative to a baseline trained on the complete dataset.  
Under nominal conditions, the \ac{mlp} classifier reaches an F1-score of~0.990, and the regressor achieves a baseline \ac{mae} of~7.799, corresponding to an average localization error of only about 8\,\% of the line length.

\begin{figure}[tbp]
    \centering
    \begin{minipage}{0.95\linewidth}
        \centering
        \includegraphics[width=\linewidth]{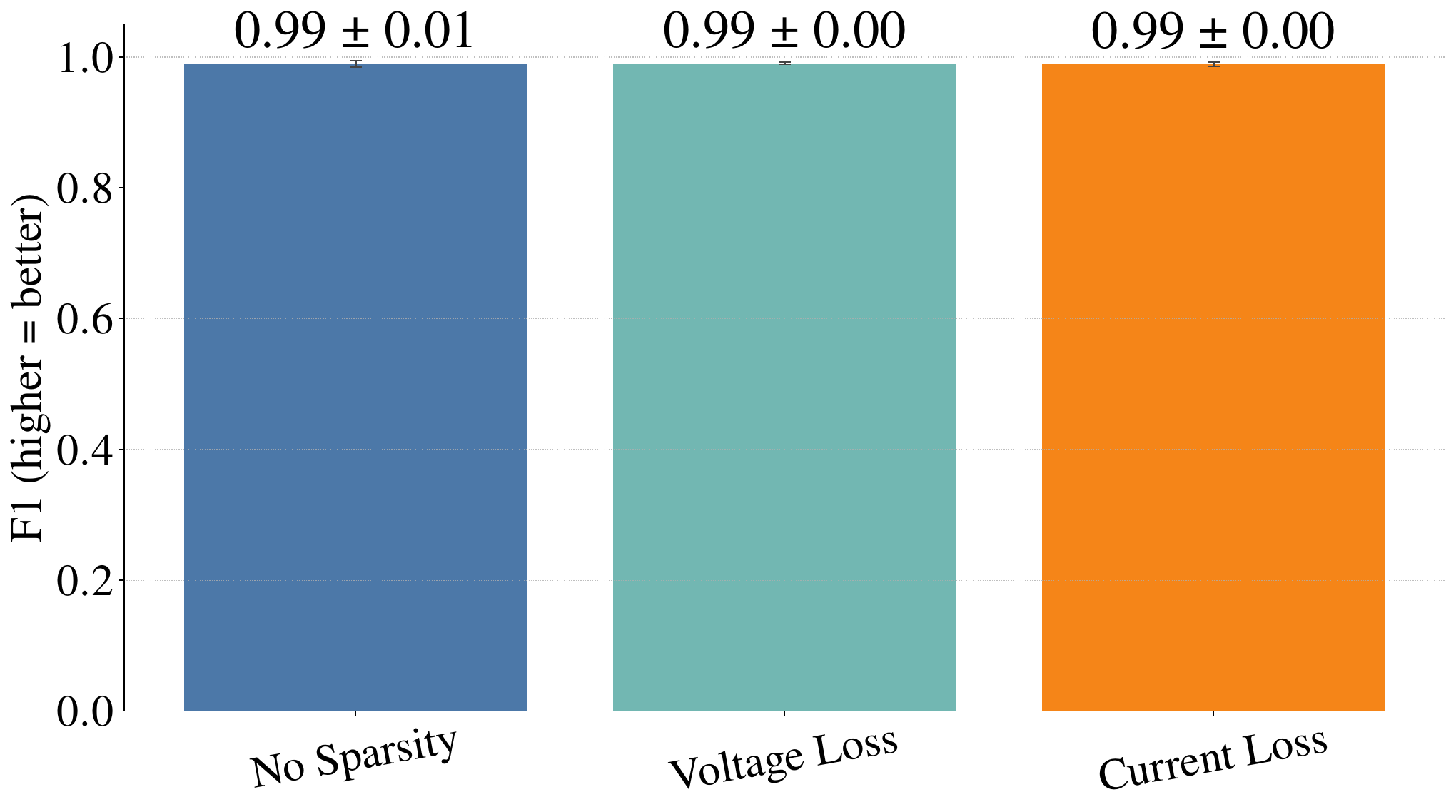}
        \subcaption{}
        \label{fig:fault_classification_f1}
    \end{minipage}
    
    \vspace{2mm}
    
    \begin{minipage}{0.95\linewidth}
        \centering
        \includegraphics[width=\linewidth]{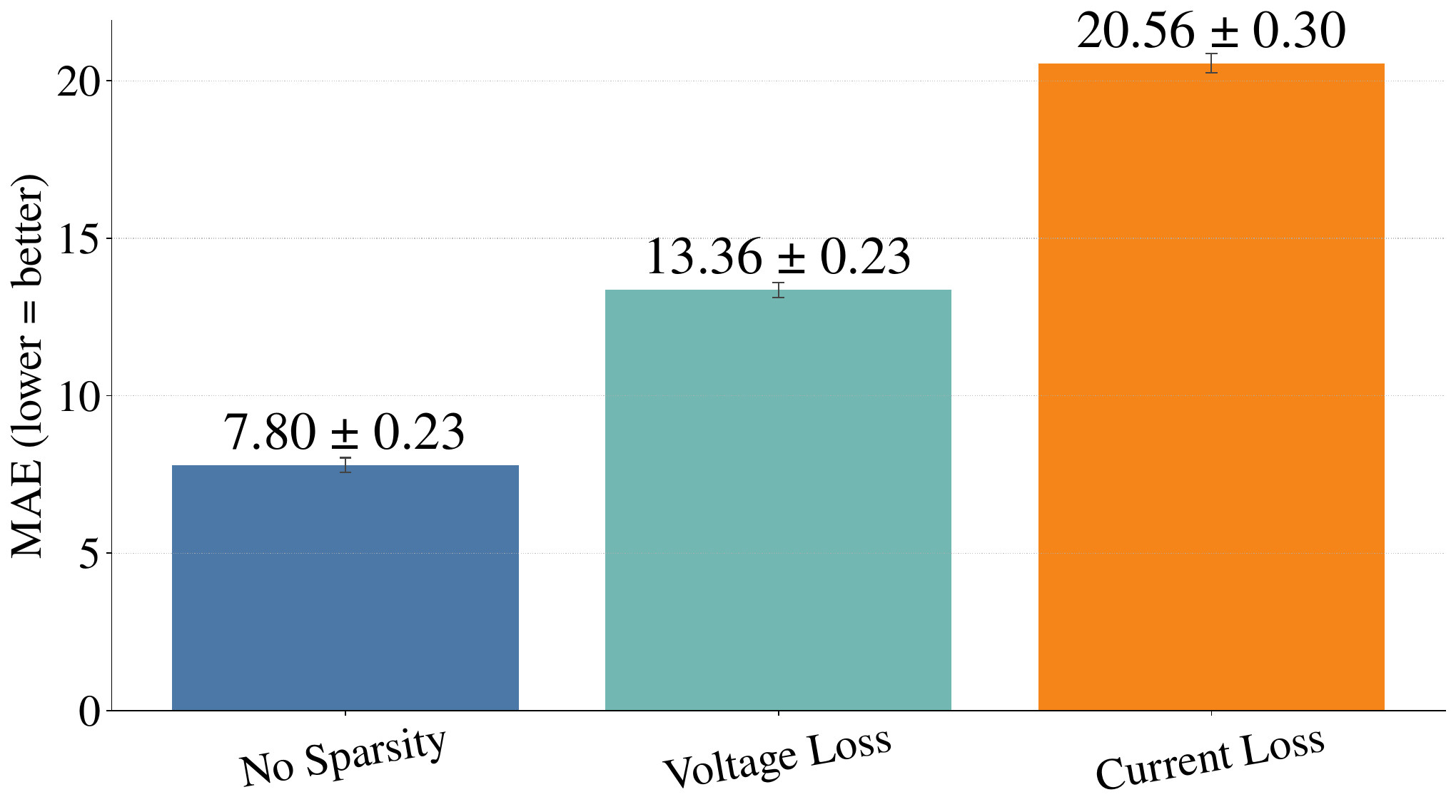}
        \subcaption{}
        \label{fig:fault_localization_mae}
    \end{minipage}
    
    \caption{Impact of current and voltage measurement loss on (a)~fault classification and (b)~fault localization performance. Bars show mean results over five-fold CV; error bars indicate the std. dev. across folds.}
    \label{fig:fc_fl_voltage_current_loss}
\end{figure}

\subsection{Current and Voltage Loss}

For \ac{fc}, the mean F1-scores remain close to 0.990 across the complete-input, voltage-loss, and current-loss conditions (Figure~\ref{fig:fault_classification_f1}). The reported relative changes are within 0.06\,\% of the complete-input reference.

In contrast, \ac{fl} is considerably more sensitive to measurement loss, as shown in Figure~\ref{fig:fault_localization_mae}. 
The \ac{mae} increases from 7.8\,\% of line length with complete inputs to 13.4\,\% under voltage loss and 20.6\,\% under current loss. These correspond to reported relative error increases of approximately 71\,\% and 163\,\%, respectively.
For the evaluated \ac{fl} model, current loss therefore causes the larger increase in localization error.
The two sensor-type ablations have much smaller effects on the reported \ac{fc} F1-scores.

\subsection{Impact of Reduced Sampling Rate}
The \ac{fc} model remains  highly robust under reduced temporal resolution.  
Up to a downsampling factor of~$\times$16 (from 6.4\,kHz to 400\,Hz), the F1-score remains nearly unaffected, staying above~0.98 with deviations below~1\% from the baseline.  
At~$\times$32 (200\,Hz), performance begins to decline more noticeably, with an F1-score of~0.96, corresponding to a~3\% reduction.  
A substantial drop is observed only at~100\,Hz, where the F1-score decreases to~0.894.  
Overall, these results indicate that reliable fault-type classification can still be achieved at moderately reduced sampling rates without significant loss of accuracy.

In contrast, the \ac{fl} task shows a gradual but steady degradation with stronger downsampling.  
The \ac{mae} remains stable around~7.7~up to~$\times$8, but increases to~8.2~at~$\times$16, representing a~5\% deviation from the baseline.  
At lower sampling frequencies, performance deteriorates more rapidly, reaching~9.75~and~14.67~for 200\,Hz and 100\,Hz, respectively.  
While moderate downsampling is acceptable, stronger reductions limit temporal detail and consequently degrade fault-distance precision.  
The summarized effects for both tasks are presented in Table~\ref{tab:downsampling_overview}.

The results suggest that \ac{fc} primarily depends on broader waveform characteristics that persist at lower resolutions, whereas \ac{fl} accuracy depends on fine temporal details of transient propagation.  
%This indicates that temporal precision, rather than overall signal shape, is the limiting factor for reliable fault distance estimation.

\begin{table}[tbp]
\centering
\caption{Impact of downsampling on \ac{fc} and \ac{fl}.}
\renewcommand{\arraystretch}{1.05}
\resizebox{0.98\linewidth}{!}{
\begin{tabular}{llcc}
\hline
\textbf{Factor} & \textbf{Freq.} & \textbf{F1~($\uparrow$)} & \textbf{\ac{mae}~($\downarrow$)} \\
\hline
No Sparsity & 6.4\,kHz & 0.989\,$\pm$\,0.005 & 7.80$\,\pm$\,0.23 \\
$\times$2 & 3.2\,kHz & 0.991\,$\pm$\,0.004 & 7.49$\,\pm$\,0.20 \\
$\times$4 & 1.6\,kHz & 0.990\,$\pm$\,0.002 & 7.61$\,\pm$\,0.15 \\
$\times$8 & 800\,Hz & 0.990\,$\pm$\,0.003 & 7.70$\,\pm$\,0.20 \\
$\times$16 & 400\,Hz & 0.985\,$\pm$\,0.001 & 8.20$\,\pm$\,0.13 \\
$\times$32 & 200\,Hz & 0.960\,$\pm$\,0.003 & 9.75$\,\pm$\,0.08 \\
$\times$64 & 100\,Hz & 0.894\,$\pm$\,0.002 & 14.67$\,\pm$\,0.15 \\
\hline
\end{tabular}}
\label{tab:downsampling_overview}
\end{table}

\subsection{Impact of Substation and Relay Communication Failures}
\begin{figure}[tbp]
    \centering
    \includegraphics[width=\linewidth]{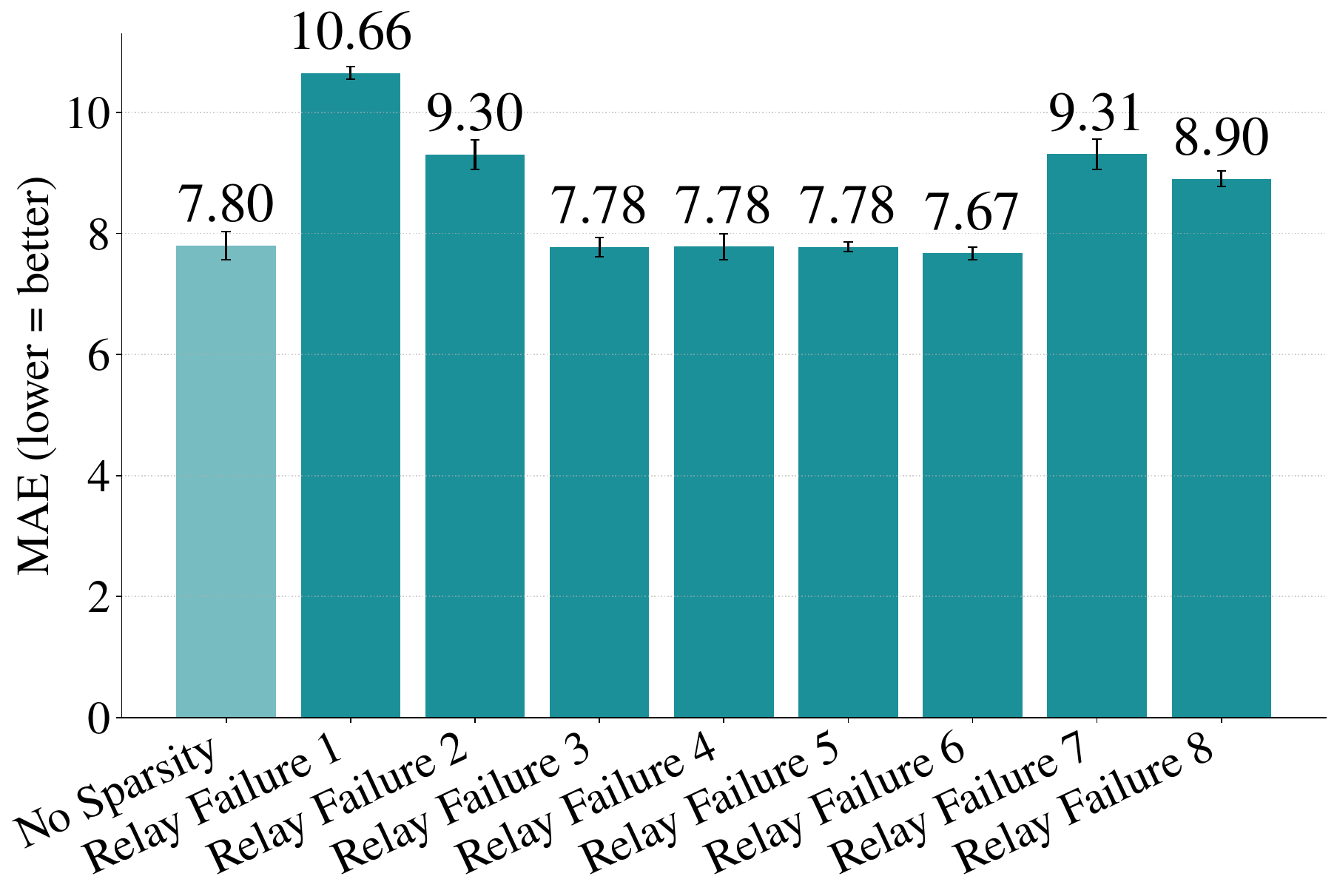}
    \caption{Impact of individual relay outages on fault localization accuracy. Bars show \ac{mae} relative to the baseline; error bars indicate the std. dev. across five-fold CV.}
    \label{fig:fault_localization_relay_failure_mae}
\end{figure}

%Relay and Substation
The \ac{fc} task remains largely unaffected by individual substation or relay outages, showing a maximum F1-score deviation of only~0.5\,\%, indicating that fault type classification relies little on global context.  
In contrast, the \ac{fl} task is considerably more sensitive: compared to the baseline \ac{mae} of~7.8, substation~1,~2, and~3 outages increase the error to~10.7,~15.9, and~9.2 rises of~37\,\%,~103\,\%, and~18\,\%, respectively.  
Relay failures have negligible impact on classification but degrade localization: as shown in Figure~\ref{fig:fault_localization_relay_failure_mae}, outages of relays 1, 2, 7, and 8 increase the \ac{mae} up to~37\,\%, while relays near substation~2 cause only minor deviations of up to $-$1.6\,\%.
The pronounced sensitivity around substation~2 suggests that the model implicitly learned spatial dependencies, where missing central measurements distort the \ac{fl} reference. In contrast, \ac{fc} appears less affected, likely relying on global disturbance patterns rather than locally available information.

\subsection{Impact of Phase Measurement Failure}
\begin{figure*}[!tb]
    \centering
    \begin{minipage}{0.46\linewidth}
        \centering
        \includegraphics[width=\linewidth]{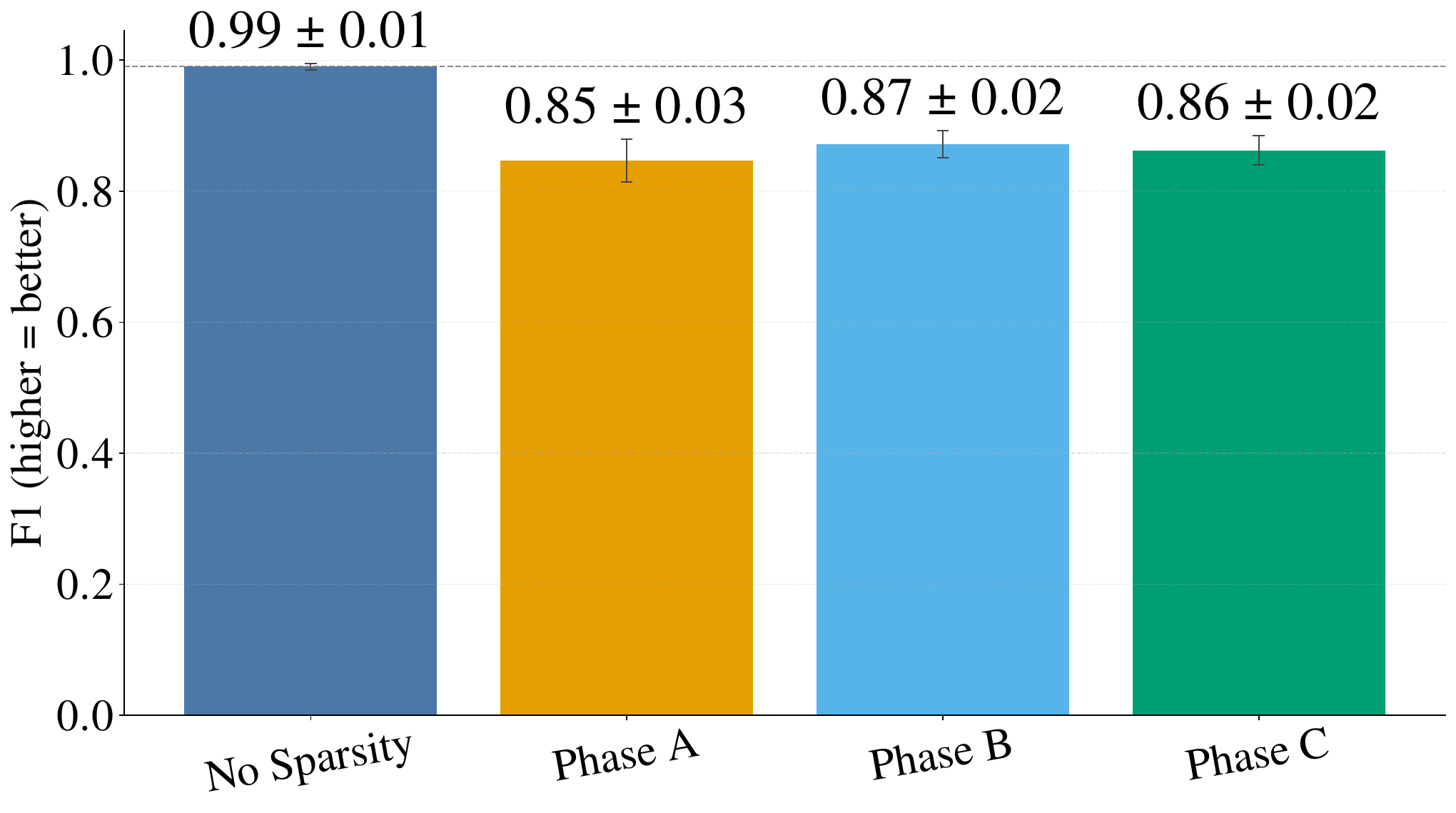}
        \subcaption{}
        \label{fig:fault_classification_phase_loss}
    \end{minipage}\hfill
    \begin{minipage}{0.46\linewidth}
        \centering
        \includegraphics[width=\linewidth]{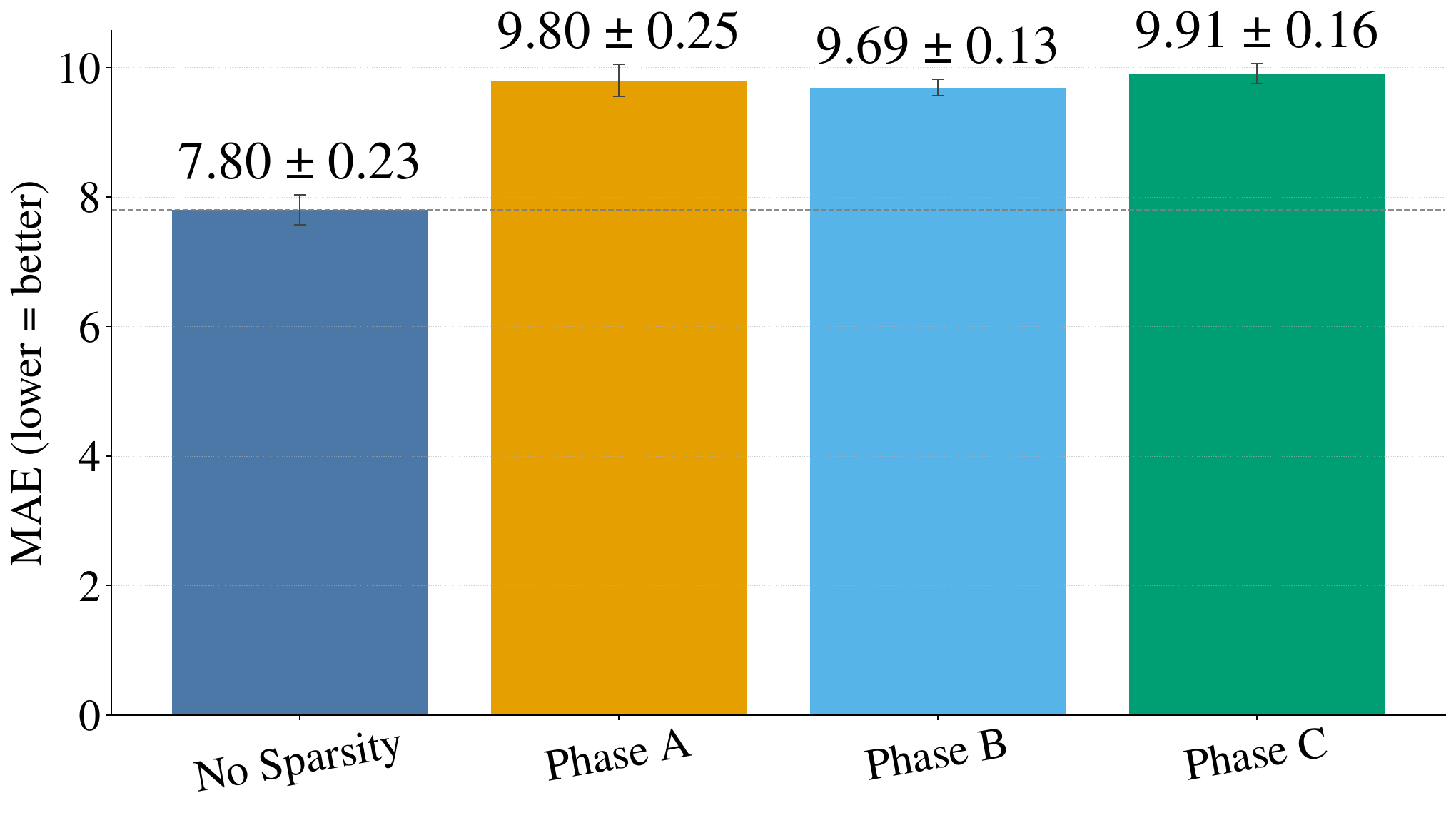}
        \subcaption{}
        \label{fig:fault_localization_phase_loss}
    \end{minipage}
    \caption{Impact of single-phase measurement loss on (a)~fault classification and (b)~fault localization performance. Bars show mean results over five-fold CV; error bars indicate the standard deviation across folds.}
    \label{fig:phase_loss_fc_fl}
\end{figure*}

Phase-loss scenarios caused the most pronounced degradation across all experiments for the \ac{fc} task. F1-scores dropping from~0.99 (baseline) to~0.85-0.87 when one phase was missing; a relative decrease of about~12-14\,\%, as in Figure~\ref{fig:fault_classification_phase_loss} 
Similarly, for the \ac{fl} task, the \ac{mae} increased from~7.8 to~ around 9.8, an average rise of roughly~26\% (see Figure~\ref{fig:fault_localization_phase_loss}).
These findings highlight that the absence of any single phase severely impairs both fault discrimination and spatial localization, emphasizing the importance of complete three-phase measurements for robust \ac{ml}-assisted protection. The observed trends are illustrated in Figure~\ref{fig:fault_classification_phase_loss}, which shows the distinct performance drop in classification under single-phase loss conditions.
The sharp drop under single-phase loss shows that the models rely on inter-phase correlations, which are disrupted when one phase is missing, impairing symmetry and spatial fault inference.

\subsection{Impact of Temporal Communication Loss}
To assess robustness against transient communication failures, contiguous 5-40\,ms segments were zeroed across all features, emulating short synchronization or communication dropouts in practical protection systems, where data streams may become temporarily unavailable.

For block durations up to~20\,ms, the \ac{fc} model maintained an almost constant F1-score between~0.986 and~0.990, deviating less than~0.5\% from the baseline.  
The \ac{fl} model even showed slightly improved localization accuracy, with the mean~\ac{mae} decreasing from~7.80~to~7.48-7.66, corresponding to relative gains of up to~4\%.  
This minor improvement is likely due to random variations in the data and not indicative of a systematic performance gain.
A measurable degradation occurs only for extended outages of~40\,ms, where the \ac{mae} increases to~8.08~(\,+3.5\%\,).  
Overall, both models show high resilience to short-term communication losses, confirming their robustness under transient data unavailability.
Short data gaps caused negligible changes in performance, whereas longer outages distort temporal dependencies; both models remain stable as long as partial context is preserved.

\section{Discussion}
The sensor-type ablations show an asymmetry between tasks: \ac{fc} scores remain close to the complete-input reference, whereas \ac{fl} has its largest error when current measurements are removed. The evaluated \ac{fc} model remains stable down to 400\,Hz, while \ac{fl} benefits from retaining inputs at 800\,Hz or above. Single-phase loss produces a marked drop in \ac{fc} performance. Short data gaps (5--20\,ms) have only minor effects on the reported scores.

The compact \ac{mlp} architecture favours interpretability and computational efficiency but underutilizes temporal and spatial dependencies. Sequence- or graph-based models, or joint \ac{fc}/\ac{fl} training, could further enhance robustness to missing data.

The authors acknowledge some limitations, including the use of a single benchmark topology, purely simulated data and a sole focus on short-circuit faults. Future work should validate transferability to other network topologies and non-fault events, as well as to real-world field conditions. It should also include additional complementary error metrics for more practical evaluation.

\section{Conclusion}
This study proposes a unified framework to evaluate the robustness of \ac{ml}-based \ac{fc} and \ac{fl} under limited observability.  
Across all scenarios, \ac{fc} proved highly stable, while \ac{fl} was more sensitive to degraded measurements.  
Among the sensor-type ablations, current loss caused the largest localization error. Substation outages and relay failures increased the \ac{mae} in several tested conditions. Classification scores remained stable down to 400\,Hz, and localization showed limited degradation down to 800\,Hz. Short communication interruptions showed negligible influence.

Despite relying on simulated data, the findings provide clear design guidance for robust, data-driven protection systems.  
They highlight the sensitivity of the evaluated localization model to loss of current channels and the need to account for communication reliability and task-specific sampling in robustness evaluations.
Future research should focus on robustness-aware training that tolerates sensor loss, the integration of physics- and topology-informed models to improve generalization, and hardware-in-the-loop validation across diverse grid configurations.  
Together, these efforts will help transform the presented framework into practical tools for reliable, ML-assisted protection in modern power systems.

\section*{Acknowledgements}
This project was funded by the Deutsche Forschungsgemeinschaft (DFG, German Research Foundation) - 535389056.

\section*{References}
\begingroup
\footnotesize
\bibliographystyle{IEEEbib}        
\bibliography{refs}         
\endgroup

\end{document}